\documentclass[conference]{IEEEtran}
\IEEEoverridecommandlockouts
\usepackage{cite}
\usepackage{amsmath,amssymb,amsfonts}
\usepackage{algorithmic}
\usepackage{graphicx}
\usepackage{textcomp}
\usepackage{xcolor}
\usepackage{graphicx}
\usepackage{comment}
\usepackage{hyperref}
\usepackage{amsmath}
\usepackage{algorithm}

\def\BibTeX{{\rm B\kern-.05em{\sc i\kern-.025em b}\kern-.08em
    T\kern-.1667em\lower.7ex\hbox{E}\kern-.125emX}}
\begin{document}

\title{ Bio-inspired Gait Imitation of Hexapod Robot Using Event-Based Vision Sensor and Spiking Neural Network\\}

\author{\IEEEauthorblockN{Justin Ting, Yan Fang, Ashwin Lele, Arijit Raychowdhury}
\IEEEauthorblockA{\textit{School of Electrical and Computer Engineering} \\
\textit{Georgia Institute of Technology}\\
Atlanta, GA, USA \\
(jting31, yan.fang, alele)@gatech.edu, arijit.raychowdhury@ece.gatech.edu}
}

\maketitle

\begin{abstract}
Learning how to walk is a sophisticated neurological task for most animals. In order to walk, the brain must synthesize multiple cortices, neural circuits, and diverse sensory inputs. Some animals, like humans, imitate surrounding individuals to speed up their learning. When humans watch their peers, visual data is processed through a visual cortex in the brain. This complex problem of imitation-based learning forms associations between visual data and muscle actuation through Central Pattern Generation (CPG). Reproducing this imitation phenomenon on low power, energy-constrained robots that are learning to walk remains challenging and unexplored. We propose a bio-inspired feed-forward approach based on neuromorphic computing and event-based vision to address the gait imitation problem. The proposed method trains a "student" hexapod to walk by watching an "expert" hexapod moving its legs. The student processes the flow of Dynamic Vision Sensor (DVS) data with a one-layer Spiking Neural Network (SNN).  The SNN of the student successfully imitates the expert within a small convergence time of ten iterations and exhibits energy efficiency at the sub-microjoule level. 

\end{abstract}


\section{Introduction}
Learning walking gaits for legged robots in real time remains a challenge due to the computational and energy constraints of battery-powered edge-computing platforms. Ideally, algorithms for learning gaits are compact and effective, using little sensing data. This approach is contradictory to popular machine learning techniques that require tremendous amount of data and computing power, such as deep learning\cite{lecun2015deep}. Instead of seeking solutions based on dense training data and power-hungry hardware, we resort to inspiration from the human brain, which is energy-efficient when learning tasks\cite{calimera2013human}. 

Learning to walk is imperative to the survival of legged vertebrates. The various motor patterns are either innate in the neural circuitry, or learned through imitation and interaction with an external environment\cite{grillner2004innate}. The ability of primates (humans in particular) to imitate others facilitates the rapid learning of new motor coordination patterns \cite{fogassi1992space,rizzolatti2001neurophysiological}. 

In this paper, we design a real-time feed-forward learning system that enables a hexapod robot to visually observe and imitate the gaits of another identical robot.  To achieve learning and energy efficiency, we leverage the combined benefits of two bio-inspired approaches: neuromorphic computing using a Spiking Neural Network (SNN) \cite{roy2019towards}, and event-based vision from a Dynamic Vision Sensor (DVS)\cite{event_survey}. Below, we briefly these concepts that are related to our work.

\textbf{Spiking Neural Networks (SNNs)} are the third generation of neural networks that model the dynamic behavior of biological neural systems \cite{maass1997networks}. SNNs encode information in either spike frequency or spike timing generated by neurons. It is capable of processing a significant amount of spatial-temporal information with a limited number of neurons and spikes\cite{ghosh2009spiking}. Recently, SNNs have been implemented on neuromorphic computing hardware to increase energy efficiency\cite{davies2018loihi, merolla2014million}. Furthermore, recent advances of emerging nanoelectronic materials and devices, such as resistive RAMs (RRAM)\cite{indiveri2013integration} and spintronic devices \cite{romera2018vowel}, are facilitating the development of real-time large-scale mixed-signal neuromorphic computing systems. These systems have the potential to bridge the energy efficiency gap between artificial systems and neural systems. SNNs have been successfully applied to various computation tasks, such as visual recognition\cite{cao2015spiking}, natural language processing\cite{diehl2016conversion}, brain-computer interface\cite{kasabov2014neucube}, and robotic control\cite{bouganis2010training}. 

\textbf{Central Pattern Generators (CPG)} are an ensemble of neural oscillators located in the spinal cord of vertebrates and in ganglions of invertebrates that are intricately involved in producing rhythmic patterns like locomotion, breathing, chewing etc\cite{steuer2019central,hooper2001central}. CPG research sheds light on the fundamental signal flow during locomotion. The dynamic behavior of CPGs can be modeled with SNNs, and thus be applied to the locomotion control of legged robots\cite{primary,fukuoka2015simple}. In this work, we configure the control pattern of each leg based on the CPG model\cite{primary,cge}.

\textbf{Dynamic Vision Sensors (DVS)} are novel vision sensors that produce a stream of asynchronous events. These events are dependent on the change in pixel intensities\cite{event_survey}. Benefiting from the decoupled pixels, DVS cameras generate "frameless" binary data at one bit per pixel. The binary property of DVS visual data shrinks the data size and allows the cameras to run in milliwatts, saving a tremendous amount of power. This level is suited for energy constrained setups\cite{event_survey}. Two additional advantages of DVS's are high dynamic range and low latency. DVS cameras operate with latencies in the range of microseconds, which is ideal for processing high-speed motion. Its wide dynamic range at a maximum of 140dB can handle scenes with large illumination disparities. Moreover, the visual data produced by DVS can be seamlessly fed into the asynchronous event-driven SNNs\cite{stromatias2017event}. The combination of DVS's and SNNs serve as an extra bonus towards reducing computational intensity and energy consumption. 

In this work, we propose a bio-inspired end-to-end real-time learning system based on SNNs and DVS's in order for a hexapod robot to perform gait imitation. In our demonstration, one hexapod robot (student) observes the gait examples of another hexapod (expert) through a DVS camera. The event data from the DVS is used to train the "student" robot, which enables the robot to imitate the "expert" robot's gait. The innovations and contributions of the proposed method are listed below:

\begin{itemize}
  \item The proposed method is compact and effective. The main components are a filter with simple AND operations, a Gaussian filter, and a one-layer SNN with six spiking neurons. The Gaussian filter only operates in the training phase. Such a compact design can be easily implemented on any edge computing platform, even without neuromorphic hardware.
  \item The proposed method is a fast real-time learning process. In our demonstration, the student hexapod robot begins to synchronize its gait into the target gait in around ten iterations of training.
  \item Our method is bio-inspired and event-driven. Taking advantage of coupling the SNN and DVS, the system is fed with a small data stream and processes the spatio-temporal information in real-time. Such a design can keep the energy consumption of the gait imitation computation in the sub-microjoule level.
  \item As far as we know, our work is the first to achieve gait imitation learning of legged robots by combining neuromorphic computing and event-based vision, both of which are bio-inspired.
\end{itemize}

The remaining parts of this paper are organized as follows. In Section II, we give a brief summary of existing related work; Section III describes the proposed method in detail. The experimental setup, results and demonstration are presented in Section IV. We also include energy estimation based on neuromorphic hardware. Section V consists of the conclusion and future work.

\section{Related Work}
Using spike-based CPG to control the hexapod robot's gait was explored in \cite{primary}. The SNN weights are obtained by utilizing a simpleton method of solving linear inequalities. Three weight matrices correspond to three different gait dynamics of CPG. Espinal et al. used an evolution algorithm to obtain weight matrices of CPG SNN \cite{cge}. Another work demonstrates an on-robot training algorithm of CPG. A reward function is formed using balance and forward translation measurements to train a spiking reinforcement learning algorithm \cite{AICAS}. However, these methods do not borrow from the natural phenomena of imitation learning. Coupling robotic motion with DVS has been done for obstacle avoidance in a non-bioinspired wheeled robot \cite{closedloop}. Movements from a video demonstration can be reproduced using a recurrently connected SNN with prediction error minimization \cite{imitation}. However, both of these methods use either a pre-programmed SNN or a non-biological learning framework. 

\begin{figure}[h!]
\centerline{\includegraphics[width=0.5\textwidth,height=0.6\textwidth]{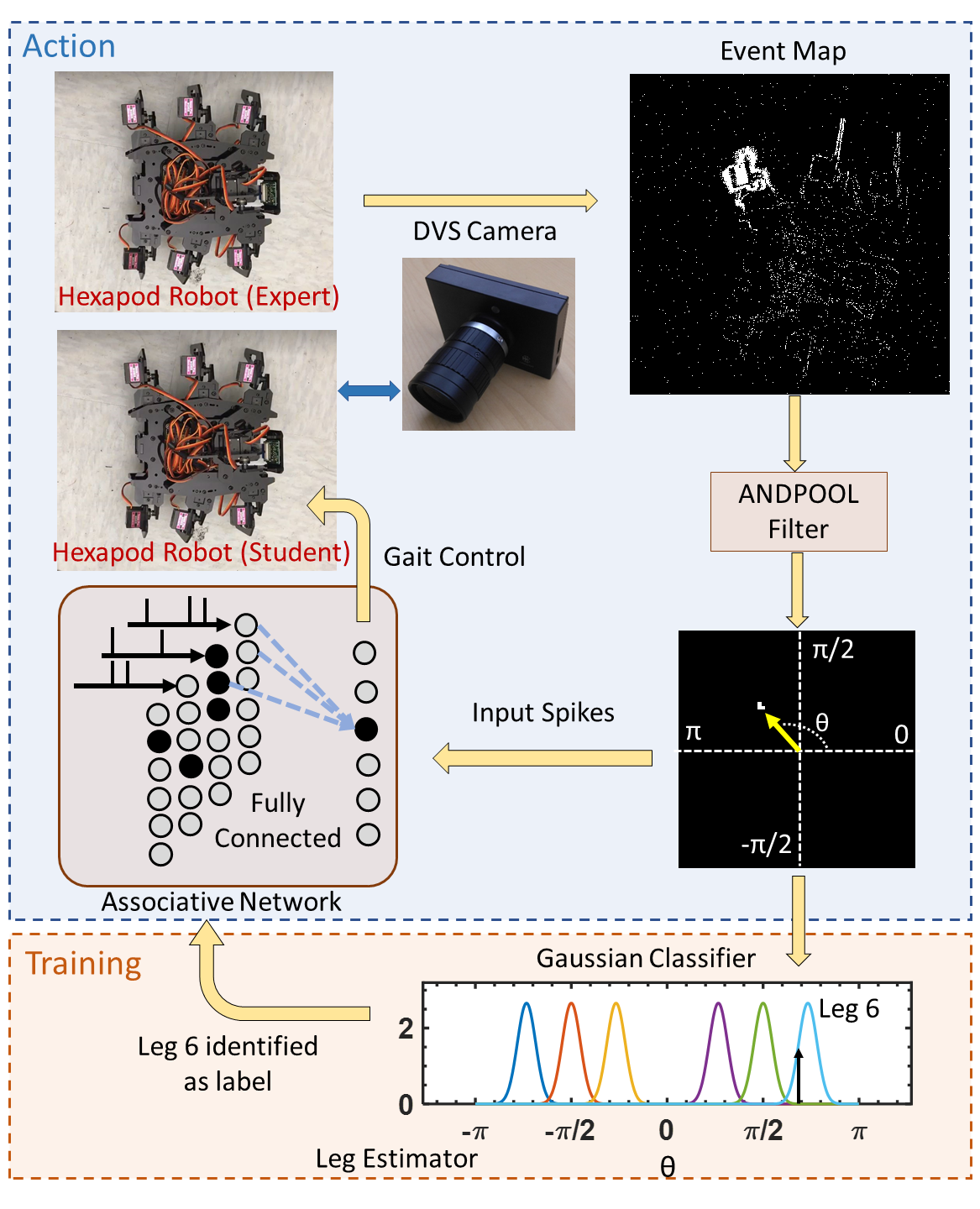}}
\centering
\caption{This figure explains the algorithm's flow. The expert hexapod moves the legs, which the DVS camera records in order to generate the event map. The event map is filtered using an Andpool to find the region of high spike generation. The relative angle of this region ($\theta$) is passed through the Gaussian classifier to generate a label for the leg that moved. The map is fed as the input layer to a six-neuron SNN. Using the label, the weights are adjusted to train the SNN. The neuron which spikes is the leg associated with the expert's leg and activated on the student hexapod.}

\label{fig:hexapodpic}
\end{figure}

\section{Method}
Fig. 1 shows the flow of data and learning architecture required for training and testing the SNN. The expert hexapod robot shown in Fig. 1 walks using CPG \cite{primary}. This walk is recorded using the DVS, which provides binary visual data. The data is filtered with a kernel for denoising and compression to identify the most active sections of the video. The SNN processes the filtered data to find associations between the legs in the video and the legs on the hexapod. During the training, the moving legs are assigned a label using a Gaussian classifier. Based on the leg labels and filtered data, the neurons in the SNN's output layer adjust their weights. The experimental setup and algorithm details are explained below.

\subsection{Event-Based Vision}\label{Events}
A CeleX5 DVS camera was used to collect event-based visual data. The pixels in the dynamic vision sensor operate asynchronously. An event is generated at a pixel if the intensity of that pixel changes. \cite{event_survey}. All the pixels operate independently. In other words, activity is only generated in the pixels where motion is detected. DVS cameras also save plenty of energy, which is especially useful for natural bio-inspired tasks that are expected to consume little energy in real animals. Quick leg movements can be efficiently captured with additional bandwidth, freeing the camera from the frames per second constraint that applies to regular frame-based cameras. Additionally, DVS cameras boast a wide dynamic range (140 dB vs. 60 dB) and reduced motion blur. 

We collect data flow from the DVS, which returns information in the form $(x,y,t)$, where $x$ and $y$ are the pixel coordinates and $t$ is the time-stamp at which the event was generated. The events generated within a 40 ms window are accumulated into a single array. We will represent this array as $I \in \{ 0,1 \} ^{n\times m}$, where $n$ and $m$ are placeholders for dimensions that will be specified later. All the data in the form $I$ will be referred to as DVS images or DVS data.

\begin{figure}[h!]
\centerline{\includegraphics [width=0.5\textwidth,height=0.3\textwidth]{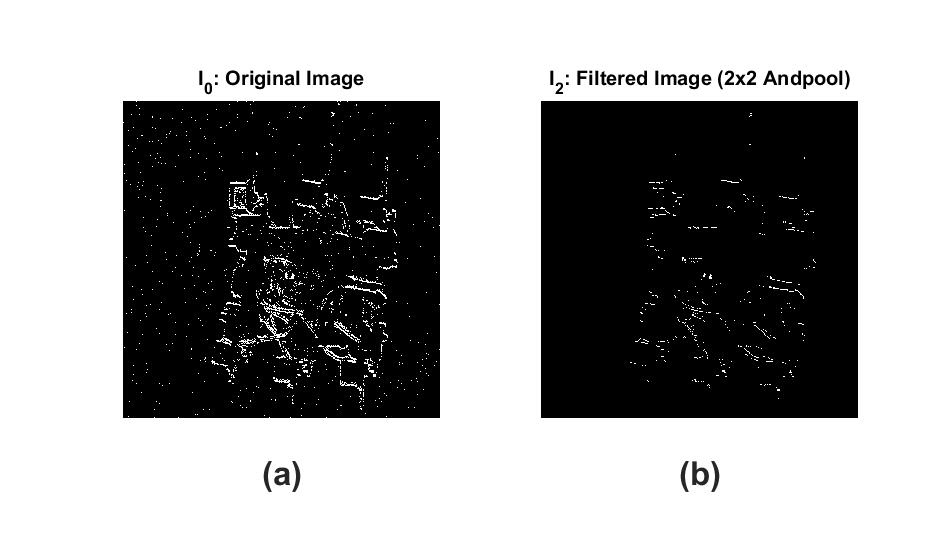}}
\centering
\caption{A comparison between a raw DVS image $I_0$ and its filtered counterpart $I_{10}$. The $2\times 2$ Andpool filter is able to remove all the noise from $I_0$. A value of 1 in any $I$ equates to a white pixel. Likewise, a 0 equates to a black pixel. The subscript of $I$ denotes how large of a filter was applied to $I_0$.}
\label{fig:filter2}
\end{figure}

 Before any of the data can be used to train the robot, a minpool filter must be applied to denoise the data, equivalent to how kernels are applied over images in Convolutional Neural Networks (CNN). Since $I \in \{ 0,1 \} ^{n\times m}$, a minpool is equivalent to an AND operation being applied over a section of the image (Fig. \ref{fig:filter2}). This AND operation over a space in  $\{ 0,1 \}^{n\times m}$ will be referred to as an Andpool in this paper. Similarly, a maxpool in $\{ 0,1 \}^{n\times m}$ is equivalent to an OR operation, so it will be referred to as an Orpool. In this paper, the subscript of $I$ will refer to how that image was filtered by an Andpool. For example, a DVS image filtered by a $10\times 10$ Andpool will be called $I_{10}$, while the original unfiltered image will be $I_0$.

\begin{figure}[h!]
\centerline{\includegraphics [trim=0 0 0 0, width=0.5\textwidth,height=.24\textwidth]{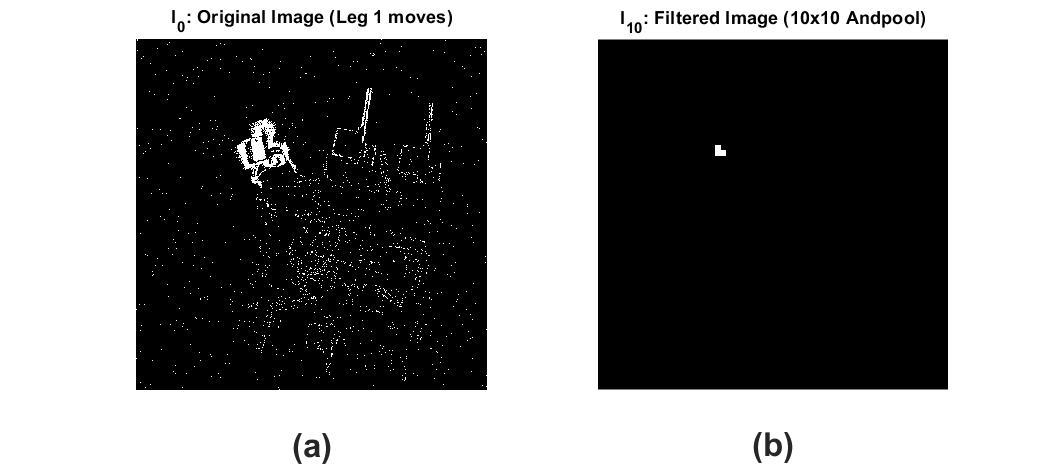}}
\centering
\caption{A comparison between $I_0$ and $I_{10}$ at the time a single leg moves. $I_0$ is a $600\times 600$ binary matrix, and $I_{10}$ is a $60\times 60$ binary matrix. In $I_{10}$, only three of the 3600 pixels are activated. In other words, the sum of $I_{10}$ at this instance of time is three. These three pixels correspond to the Leg 1. Fig. \ref{fig:leg_spikes} displays the $I_{10}$ for each of the legs.}
\label{fig:I10}
\end{figure}

\begin{figure}[h!]
\centerline{\includegraphics[width=0.5\textwidth,height=0.35\textwidth]{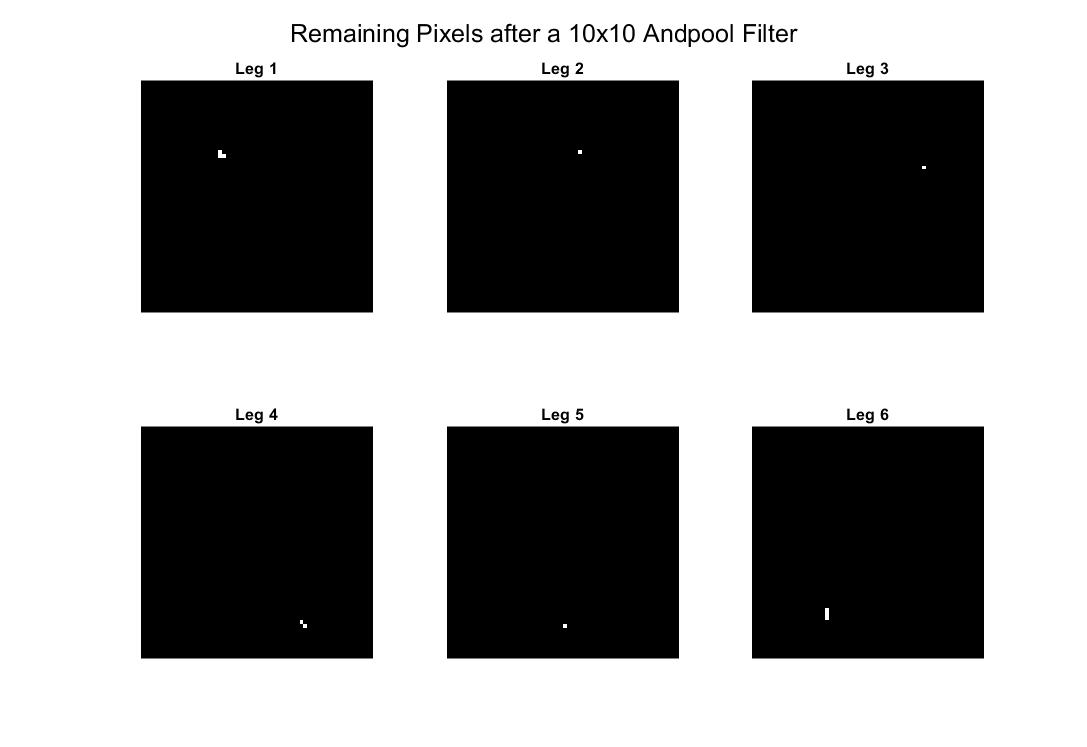}}
\centering
\caption{Each one of the hexapod's legs are visible in $I_{10}$ if all the legs move one at a time in the video. $I_{10}$ is noticeably sparse, which makes it easy for the algorithm to determine not only which leg moved, but also when it moved. These frames correspond to the peaks in Fig. \ref{fig:spikechart}(b).}
\label{fig:leg_spikes}
\end{figure}

\begin{figure}[h!]
\centerline{\includegraphics[width=0.5\textwidth,height=0.4\textwidth]{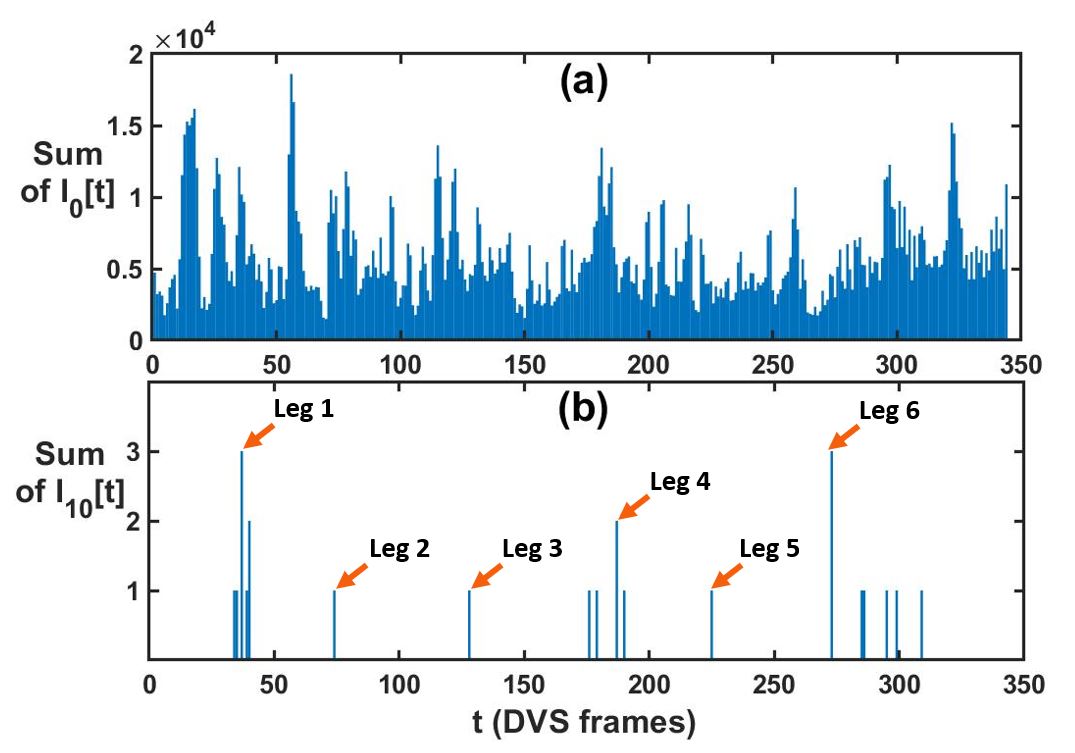}}
\centering
\caption{A comparison of the number of white pixels per frame between $I_0$ and $I_{10}$. Once again, the sparseness of $I_{10}$ is apparent. The Andpool's ability to accurately detect leg movement helps simplify the algorithm for the associative network. The filter is so exclusive that Legs 2, 3, and 5 only have one pixel to represent them.}
\label{fig:spikechart}
\end{figure}

\subsection{SNN Architecture}\label{single_neuron}
An SNN architecture was used to train the robot. Both DVS and SNN use a  continuous flow of binary events, in which the presence of a spike at a particular time is equivalent to a value of $1$ in an array, and $0$ otherwise. Therefore, the coupling of these two bio-inspired systems is natural and intuitive. We feed a stream of DVS data into the SNN, eliminating the need for time stamps and reducing memory consumption. The many-input, single-output neuron model obeys Equation (1). If the sum of inputs increases, the action potential $V$ of the neuron increases. If $V$ reaches a threshold, the neuron "spikes", which is equivalent to saying that the neuron's output at a particular time is $1$. Otherwise, the neuron's output is $0$.

\begin{figure}[h!]
\centerline{\includegraphics[width=0.5\textwidth,height=0.45\textwidth]{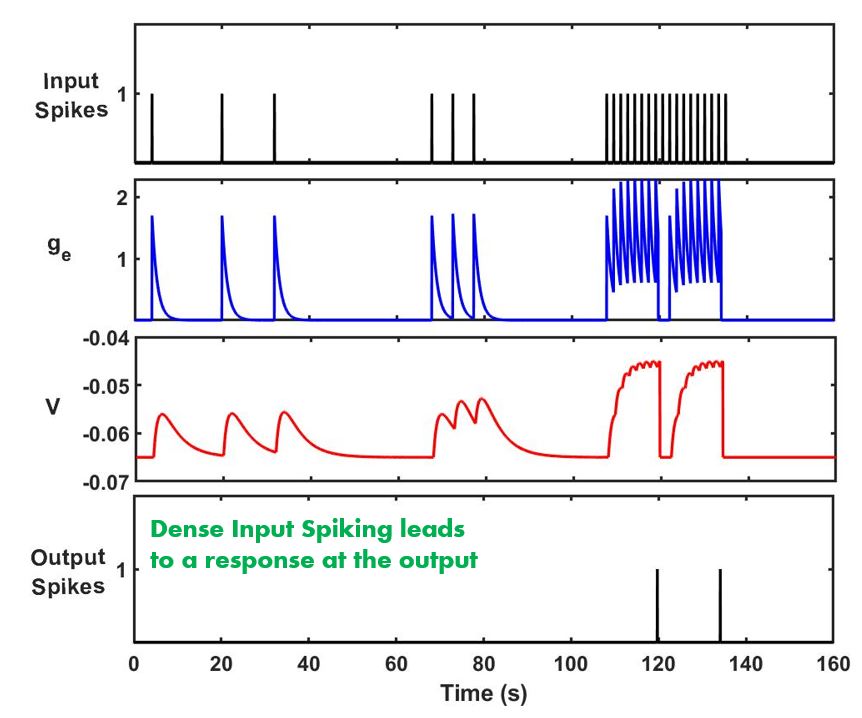}}
\centering
\caption{Demonstration of how a single neuron's output is a function of its input. The dot product between the input spikes and weights is added asynchronously to $g_e$ (Equation 3), which increases $V$. The first six input spikes are too sparse to generate an output spike, but the train of spikes that starts after 100 seconds is concentrated enough to push the action potential $V$ past the spiking threshold, resulting in two output spikes. A refractory period (2s where $V=V_{rest}$) follows the output spike. In this particular example there is only one input, but a neuron in an SNN will have multiple inputs that will be summed and added to $g_e$.}
\label{fig:neuron}
\end{figure}

A visual interpretation for the dynamics of the neuron are described by the following equations and Table \ref{tab:variables}. Fig. \ref{fig:neuron} provides a visual representation.
\begin{equation}\label{dV/dt}
\tau \frac{dV}{dt} = (V-Vrest)+g_e(V-Vexec)
\end{equation}
\begin{equation}\label{dg/dt}
\tau_{g_e} \frac{dg_e}{dt} = -g_e
\end{equation}
\begin{equation}\label{ge}
g_e = g_e + w \cdot I
\end{equation}

\begin{table}
\caption {Variables} 
\label{tab:variables}
\begin{center}
\begin{tabular}{| c | c | c |}
\hline
$\tau$ & neuron time constant, $\approx 4s$\\
\hline
$\tau_{g_e}$ & $g_{e}$ time constant, $\approx 1.2s$\\
\hline
$V_{rest}$ & resting voltage,  $\approx -65mV$\\
\hline
$V_{exec}$ & exciting voltage,  $\approx -30mV$\\
\hline
$w \in \mathbb{R}^{n\times m}$ & neuron's weights\\
\hline
$I \in \{0, 1\}^{n\times m}$ & single frame from video\\
\hline
\end{tabular}
\end{center}
\end{table}

The network consisting of these neurons is shown in Fig. 1. Six output neurons corresponding to the six legs of the student hexapod form the output layer. The training data that is fed to these output neurons consists of one filtered DVS video that repeats itself infinitely. The DVS data at time $t$ is filtered to generate a 60 $\times$ 60 image ($I_{10}[t]$). An output neuron spike is triggered when a sufficient number of input spikes in $I_{10}$ are accumulated. This trigger moves the corresponding leg on the student hexapod. Thus, training this network involves adjusting the weights connecting the input to the output layer, which strengthens one-to-one associations the legs and $I_{10}$. The output is fully connected to the input, and the weights are initialized to zero at the beginning of training. Equations \ref{dg/dt} and \ref{ge} illustrate the behavior of input synapses, which are incorporated in the action potential dynamics of spiking neurons.

\subsection{Weight Adjusting Algorithm}\label{sub:weight_change}
The algorithm for adjusting weights involves associating leg movement in the training video ($I_0[t]$) to the leg on the student hexapod. First, to clear the random noise, the data is filtered with a $10\times 10$ Andpool to form $I_{10}$. The existence of any spikes at all in $I_{10}$ indicates that a leg moves. The leg label, which is an integer between 1 and 6, associated with those spikes is identified using the Gaussian classifier described in Section \ref{LegEst} and Algorithm \ref{alg:leg_estimate}. This label is used as an index for the output layer. The weights that are connected to the spikes are incremented by an arbitrary amount $p$, where $$p = \omega e^{-\alpha t}$$ with $\alpha, \omega$ as arbitrary constants. Since $p\rightarrow 0$ as $t\rightarrow \infty $, the SNN eventually stops training with large enough $t$. The algorithm also dampens the connections which do not receive any spikes. The end result is a feed-forward network that builds associations between areas in the image to one of the six neurons. The pseudocode is provided in Algorithm \ref{alg:correction}.

\begin{algorithm}
\caption{Weight Adjusting Algorithm}
\begin{algorithmic}[1]
\label{alg:correction}
\STATE $N \gets$ array of 6 spiking neurons
\STATE $\alpha, \beta,\omega, \sigma \gets$ constants
\STATE $u,v \gets$ length and width of frame in $I_{10}$
\FOR {each neuron in N}
\STATE $neuron.weights \gets 0^{u,v}$
\ENDFOR

\FOR {$t = 1$ to $T$}
\STATE $p \gets \omega e^{-\alpha t}$
\STATE $m \gets \sigma e^{-\beta t}$
\newline
\FOR {each neuron in N}
\STATE Update and evaluate neuron as described in equations 1-3, where the input is $I_{10}[t]$
\ENDFOR
\newline

\IF {$any(I_{10}[t])$}
\STATE $leg \gets$ Algorithm \ref{alg:leg_estimate}
\STATE$i_p \gets$ inidices where $I_{10}[t] == 1$
\STATE$i_m \gets$ inidices where $I_{10}[t]==0$
\STATE$N(leg).weights[i_p] \mathrel{+}= p$
\STATE$N(leg).weights[i_m] \mathrel{-}= m $
\ENDIF
\ENDFOR
\end{algorithmic}
\end{algorithm}

\subsection{Leg Estimator}\label{LegEst}
The solution to figuring out exactly which leg is moving at time $t$ is non-unique. Either of the front or back legs could be labeled as Leg 1. In other words, four of the six images in Fig. \ref{fig:leg_spikes} could be assigned to Leg 1, and the outcome would be the same. Therefore, we must impose our own interpretation of how the legs are numbered.

A probabilistic Gaussian classifier assigns a label to each leg action detected in $I_{10}$. From $I_{10}$, we extrapolate the leg's position which we will refer to as $l \in \mathbb{Z}^{2 \times 1}$. In order to extrapolate the hexapod's body, we must use the smaller $2\times 2$ Andpool. After the $2\times 2$ filter we follow it with a $10\times 10$ Orpool to enhance the remaining pixels. The result is more accurate if this filter is performed in the interval $[t-10, t]$, as described in step 2 of Algorithm \ref{alg:leg_estimate}. We will refer to this final result as $I_{body}$. The centroid of $I_{body}$ will be called $c \in \mathbb{Z}^{2 \times 1}$. 
The angle $\theta$ between $l$ and $c$ is used to determine the leg label, which is calculated using step 17 of Algorithm \ref{alg:leg_estimate}.
Using $\theta$, a number 1-6 can be associated with each of the frames in $I_{10}$ by taking the leg with the highest value as predicted by the Gaussian classifier in Fig. \ref{fig:gauss}.

This part of the algorithm is only used in the training. When a different gait is tested, the SNN already has the connections needed to walk correctly.

\begin{algorithm}
\caption{Leg Estimation Algorithm}
\label{alg:leg_estimate}
\begin{algorithmic}[1]
\STATE $I_{body} \gets 0^{u \times v}$ ($u$ and $v$ same as Algorithm \ref{alg:correction})
\FOR{i = 0 to 10}
\STATE $I_a \gets I_2[t-i]$
\STATE $I_b \gets$ filter $I_a$ with $10\times 10$ Orpool
\STATE $I_{body} \gets I_{body} \lor I_b$
\ENDFOR
\STATE $I_{body} \gets \lnot I_{10}[t] \land I_{body}$
\STATE $c \gets centroid (I_{body})$
\STATE $l \gets centroid (I_{10}[t])$
\STATE $\theta \gets \arctan2(c_y-l_y, c_x-l_x)$
\STATE $leg \gets \underset{L}{\arg \max} P(L|\Theta = \theta)$ (from Fig. \ref{fig:gauss})
\STATE return $leg$
\end{algorithmic}
\end{algorithm}

\begin{figure}[h!]
\centerline{\includegraphics[width=0.56\textwidth,height=0.2\textwidth]{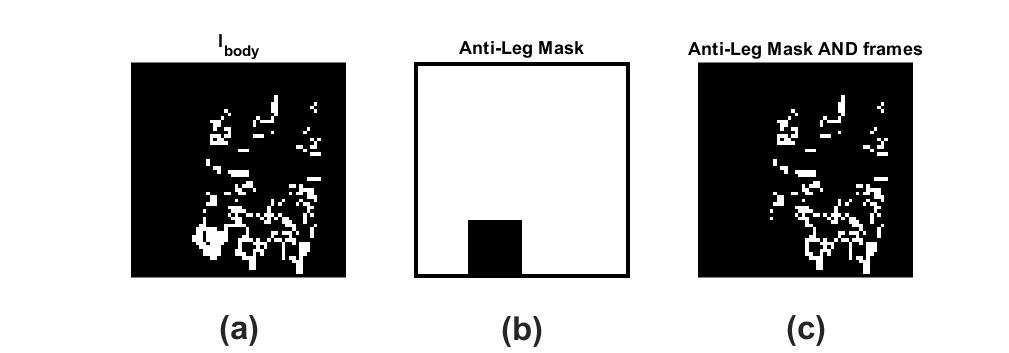}}
\centering
\caption{Visual representation of $I_{body}$, the Anti-Leg Mask, and the bitwise AND operation applied between the two. The Anti-Leg Mask removes the concentration of pixels at the leg from $I_{body}$, so that only the body is left in (c). Steps 3, 4, and 5 in Algorithm \ref{alg:leg_estimate} correspond to (a), (b), and (c). The centroid $c$ can be found from (c), whereas the leg position $l$ can be found from $I_{10}$}
\label{fig:mask}
\end{figure}

\begin{figure}[h!]
% \centerline{\includegraphics[width=0.5\textwidth,height=0.17\textwidth]{{"Figures 2"}.png}}
\centerline{\includegraphics[width=0.5\textwidth,height=0.18\textwidth]{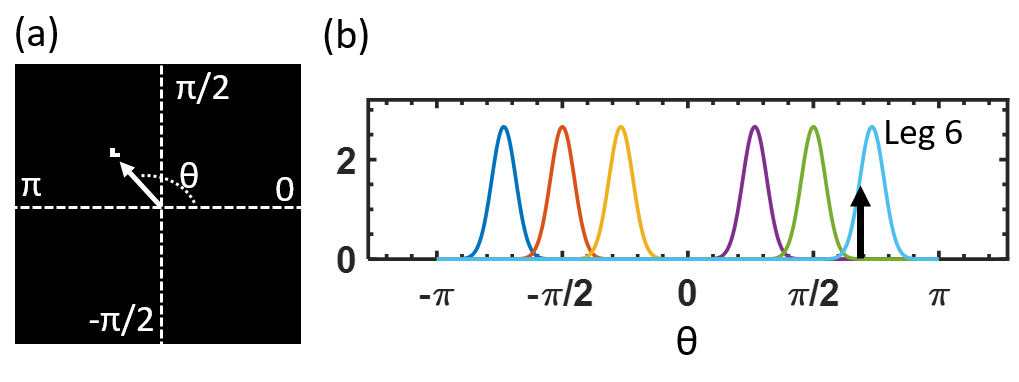}}
\centering
\caption{The picture (a) is a frame of $I_{10}$, and (b) plots the prior Gaussians for each leg. The white arrow in (a) starts at the origin $c$ and points to the group of pixels $l$. The angle $\theta$ can be calculated by taking the $\arctan$ between the coordinates of $c$ and $l$. Once $\theta$ is obtained, we can compute the distribution $P(L|\Theta = \theta)$. The leg label is the $\arg \max$ of this distribution. (a) and (b) correspond with steps 10 and 11 in Algorithm \ref{alg:correction}, respectively.}
\label{fig:gauss}
\end{figure}

\section{Results}
\subsection{Experiment Configuration}
Two hexapod robots are used as the student and the expert. A Raspberry Pi 3 on the hexapod controls its 12 servos. Each leg has two servos, allowing two DOF per leg. The legs are programmed to oscillate at the same frequency, but with phase differences between the legs. This simple method of programming a gait into the robot is inspired by CPG \cite{cge}. In order to translate the SNN output to CPG, a leg goes through one period of an oscillation when its corresponding neuron spikes. It must wait for another spike from the neuron to continue to the next cycle. Fig. 9 demonstrates how spikes translate to gaits. 

For the video recording, the robot was programmed to move one leg at a time, and the next leg does not start moving until the previous leg finishes its cycle. Once all the legs finish their cycles, the video ends. \href{https://youtu.be/_zQ1Hy_kIt8}{\underline{\color{blue}Video-1}} shows the input video, as well as the filtering steps before it is fed into the SNN. 

In order to create the new video for testing the SNN on different gaits, we actually took the original training video and cut it into six pieces, one for each leg. Then the pieces are rearranged in a different order. If two or more legs move at the same time, then the two pieces are superimposed on each other using bitwise OR. Therefore, the hexapod can be controlled by manipulating the video it was trained on.

The DVS camera used to record the video is fabricated by CelePixel Technology, which comes with a dynamic range of 140dB and a resolution of 1280$\times$800.
Each frame in the DVS video approximately equates to 0.04 seconds passing in real time. The dynamic behavior of the neuron was also configured to have 0.04 seconds pass with each iteration. The Forward Euler method was used to model Equations \ref{dV/dt}-\ref{ge}.

\subsection{Learning}

Fig. 9 shows the SNN converging to the correct solution. A demonstration is provided in \href{https://youtu.be/lg8M2_2aTFc}{\underline{\color{blue}Video-2}}. Fig. 9(a) plots the action potential build-up of one neuron in the training phase. The action potential settles down to regular periodic spiking as $t \rightarrow \infty$. Fig. 9(c) shows spiking of all six output neurons using a raster plot. The training begins by with only some of output neurons firing correctly. The pattern finally converges to the pattern in the video within 120 seconds. This completes the association between the video and the hexapod legs. 
After the training, any rearrangement of the order in which the legs move can be reproduced by the SNN. In Fig. 9(b) and 9(d), a stable tripod gait pattern (with odd and even number neurons spiking in adjacent time steps) is produced by the SNN by rearranging the input accordingly. If this network is deployed on the hexapod, the hexapod will demonstrate the tripod gait as observed in the video.

\subsection{Energy Consumption Estimation}
We estimate the computing energy consumption of spiking neurons in the testing phase. Most of the power is consumed by the Andpool filter and the SNN. 
The estimations assume implementation on customized neuromorphic computing hardware, such as Intel's Loihi \cite{davies2018loihi}, which costs 1.7 nJ per event spike. The input fires 1$\sim$3 spikes during the movement of of a single leg, as shown in Fig. 5(b), resulting in a total energy cost of around 5nJ.
For the Andpool operation, 3600 kernels of size $10\times 10$ are applied over a $600\times 600$ array. This operation can be computed with either neuromorphic systems, or traditional digital systems such as FPGA and ASIC. In the latter scenario, we assume 3600 AND gates that consist of six transistors each in 14nm technology. 
The consumption in a single transistor is in the fJ-level according to $\frac{1}{2} C V^2$. The Boolean computing of the Andpool filter costs approximately $2\sim3$ nJ. Therefore, the total computational energy consumption of processing one leg is 10nJ. Considering the other peripheral circuits and the buffering of sparse event data, a conservative estimation of the system energy cost is at the sub-$\mu$J level. Our design is extremely useful in energy constrained circumstances, such as dealing with a limited supply of battery power. The system's energy efficiency is the result of two factors. The first factor is the small and sparse visual data flow generated by the DVS, followed by the simple Boolean Andpool filter. The other factor is the asynchronous event-driven visual processing of the SNN, which conserves a significant amount of energy between input spikes.

\begin{figure}[h!]
\centerline{\includegraphics[width=0.5\textwidth,height=0.64\textwidth]{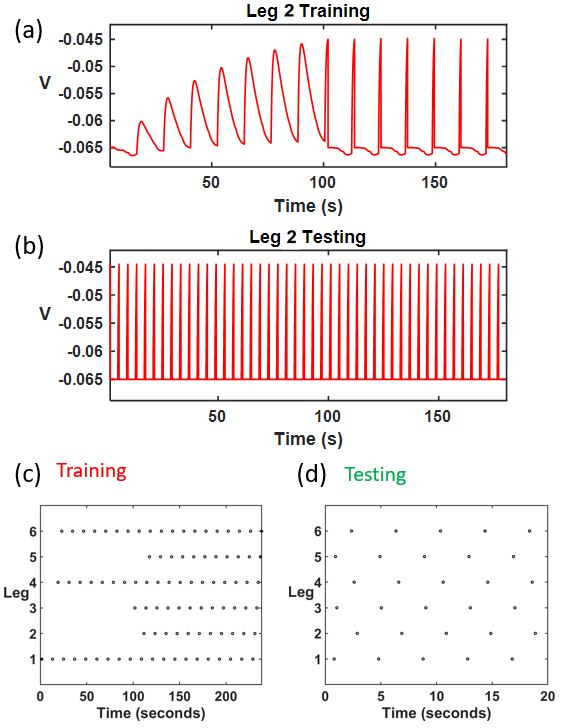}}
\centering
\caption{A comparison of action potentials ((a) and (b)) and raster plots ((c) and (d)) before and after the SNN is trained. For the action potential plots, we chose to plot the neuron corresponding to Leg 2. For the raster plots, all six neurons are displayed. In (a) and (c), the SNN is initialized with all weights are 0, but the relevant weights increase until it converges to a solution in about 100 seconds. At this time, the robot with the SNN is synchronized with the target robot. For (b) and (d), the SNN picks up a new walking pattern without any re-training. Note that the time axes between training and testing are scaled differently.}
\label{fig:train_gait}
\end{figure}

\section{Conclusion}

We proposed a bio-inspired feed-forward learning system that trains a student hexapod to imitate the gait of an expert hexapod. We exploit the inherent coupling between DVS and SNN to generate and process event-based data. The student hexapod generates the sequence of leg motions identical to the expert's by watching the expert in real time. Furthermore, the student learns the gait within a short training period, while using only one video as a data source. Ultra-low energy consumption in the sub-microjoule region makes this work competent in energy constrained scenarios and hardware platforms. Our future work will focus on real-time imitation learning that extends over more complex actions involving different state and action spaces. 

\bibliographystyle{ieeetr}
\bibliography{main}

\begin{thebibliography}{10}

\bibitem{lecun2015deep}
Y.~LeCun, Y.~Bengio, and G.~Hinton, ``Deep learning,'' {\em nature}, vol.~521,
  no.~7553, pp.~436--444, 2015.

\bibitem{calimera2013human}
A.~Calimera, E.~Macii, and M.~Poncino, ``The human brain project and
  neuromorphic computing,'' {\em Functional neurology}, vol.~28, no.~3, p.~191,
  2013.

\bibitem{grillner2004innate}
S.~Grillner and P.~Wall{\'e}n, ``Innate versus learned movements—a false
  dichotomy?,'' in {\em Progress in Brain Research}, vol.~143, pp.~1--12,
  Elsevier, 2004.

\bibitem{fogassi1992space}
L.~Fogassi, V.~Gallese, G.~Di~Pellegrino, L.~Fadiga, M.~Gentilucci, G.~Luppino,
  M.~Matelli, A.~Pedotti, and G.~Rizzolatti, ``Space coding by premotor
  cortex,'' {\em Experimental Brain Research}, vol.~89, no.~3, pp.~686--690,
  1992.

\bibitem{rizzolatti2001neurophysiological}
G.~Rizzolatti, L.~Fogassi, and V.~Gallese, ``Neurophysiological mechanisms
  underlying the understanding and imitation of action,'' {\em Nature reviews
  neuroscience}, vol.~2, no.~9, pp.~661--670, 2001.

\bibitem{roy2019towards}
K.~Roy, A.~Jaiswal, and P.~Panda, ``Towards spike-based machine intelligence
  with neuromorphic computing,'' {\em Nature}, vol.~575, no.~7784,
  pp.~607--617, 2019.

\bibitem{event_survey}
G.~Gallego, T.~Delbruck, G.~Orchard, C.~Bartolozzi, B.~Taba, A.~Censi,
  S.~Leutenegger, A.~Davison, J.~Conradt, K.~Daniilidis, {\em et~al.},
  ``Event-based vision: A survey,'' {\em arXiv preprint arXiv:1904.08405},
  2019.

\bibitem{maass1997networks}
W.~Maass, ``Networks of spiking neurons: the third generation of neural network
  models,'' {\em Neural networks}, vol.~10, no.~9, pp.~1659--1671, 1997.

\bibitem{ghosh2009spiking}
S.~Ghosh-Dastidar and H.~Adeli, ``Spiking neural networks,'' {\em International
  journal of neural systems}, vol.~19, no.~04, pp.~295--308, 2009.

\bibitem{davies2018loihi}
M.~Davies, N.~Srinivasa, T.-H. Lin, G.~Chinya, Y.~Cao, S.~H. Choday, G.~Dimou,
  P.~Joshi, N.~Imam, S.~Jain, {\em et~al.}, ``Loihi: A neuromorphic manycore
  processor with on-chip learning,'' {\em IEEE Micro}, vol.~38, no.~1,
  pp.~82--99, 2018.

\bibitem{merolla2014million}
P.~A. Merolla, J.~V. Arthur, R.~Alvarez-Icaza, A.~S. Cassidy, J.~Sawada,
  F.~Akopyan, B.~L. Jackson, N.~Imam, C.~Guo, Y.~Nakamura, {\em et~al.}, ``A
  million spiking-neuron integrated circuit with a scalable communication
  network and interface,'' {\em Science}, vol.~345, no.~6197, pp.~668--673,
  2014.

\bibitem{indiveri2013integration}
G.~Indiveri, B.~Linares-Barranco, R.~Legenstein, G.~Deligeorgis, and
  T.~Prodromakis, ``Integration of nanoscale memristor synapses in neuromorphic
  computing architectures,'' {\em Nanotechnology}, vol.~24, no.~38, p.~384010,
  2013.

\bibitem{romera2018vowel}
M.~Romera, P.~Talatchian, S.~Tsunegi, F.~A. Araujo, V.~Cros, P.~Bortolotti,
  J.~Trastoy, K.~Yakushiji, A.~Fukushima, H.~Kubota, {\em et~al.}, ``Vowel
  recognition with four coupled spin-torque nano-oscillators,'' {\em Nature},
  vol.~563, no.~7730, pp.~230--234, 2018.

\bibitem{cao2015spiking}
Y.~Cao, Y.~Chen, and D.~Khosla, ``Spiking deep convolutional neural networks
  for energy-efficient object recognition,'' {\em International Journal of
  Computer Vision}, vol.~113, no.~1, pp.~54--66, 2015.

\bibitem{diehl2016conversion}
P.~U. Diehl, G.~Zarrella, A.~Cassidy, B.~U. Pedroni, and E.~Neftci,
  ``Conversion of artificial recurrent neural networks to spiking neural
  networks for low-power neuromorphic hardware,'' in {\em 2016 IEEE
  International Conference on Rebooting Computing (ICRC)}, pp.~1--8, IEEE,
  2016.

\bibitem{kasabov2014neucube}
N.~K. Kasabov, ``Neucube: A spiking neural network architecture for mapping,
  learning and understanding of spatio-temporal brain data,'' {\em Neural
  Networks}, vol.~52, pp.~62--76, 2014.

\bibitem{bouganis2010training}
A.~Bouganis and M.~Shanahan, ``Training a spiking neural network to control a
  4-dof robotic arm based on spike timing-dependent plasticity,'' in {\em The
  2010 International Joint Conference on Neural Networks (IJCNN)}, pp.~1--8,
  IEEE, 2010.

\bibitem{steuer2019central}
I.~Steuer and P.~A. Guertin, ``Central pattern generators in the brainstem and
  spinal cord: an overview of basic principles, similarities and differences,''
  {\em Reviews in the Neurosciences}, vol.~30, no.~2, pp.~107--164, 2019.

\bibitem{hooper2001central}
S.~L. Hooper, ``Central pattern generators,'' {\em e LS}, 2001.

\bibitem{primary}
H.~Rostro-Gonzalez, P.~A. Cerna-Garcia, G.~Trejo-Caballero, C.~H.
  Garcia-Capulin, M.~A. Ibarra-Manzano, J.~G. Avina-Cervantes, and
  C.~Torres-Huitzil, ``A cpg system based on spiking neurons for hexapod robot
  locomotion,'' {\em Neurocomputing}, vol.~170, pp.~47--54, 2015.

\bibitem{fukuoka2015simple}
Y.~Fukuoka, Y.~Habu, and T.~Fukui, ``A simple rule for quadrupedal gait
  generation determined by leg loading feedback: a modeling study,'' {\em
  Scientific reports}, vol.~5, p.~8169, 2015.

\bibitem{cge}
A.~Espinal, H.~Rostro-Gonzalez, M.~Carpio, E.~I. Guerra-Hernandez,
  M.~Ornelas-Rodriguez, and M.~Sotelo-Figueroa, ``Design of spiking central
  pattern generators for multiple locomotion gaits in hexapod robots by
  christiansen grammar evolution,'' {\em Frontiers in neurorobotics}, vol.~10,
  p.~6, 2016.

\bibitem{stromatias2017event}
E.~Stromatias, M.~Soto, T.~Serrano-Gotarredona, and B.~Linares-Barranco, ``An
  event-driven classifier for spiking neural networks fed with synthetic or
  dynamic vision sensor data,'' {\em Frontiers in neuroscience}, vol.~11,
  p.~350, 2017.

\bibitem{AICAS}
A.~S. LeLe, Y.~Fang, J.~Ting, and A.~Raychowdhury, ``Learning to walk: Spike
  based reinforcement learning for hexapod robot central pattern generation,''
  in {\em 2019 IEEE International Conference on Artificial Intelligence
  Circuits and Systems (AICAS)}, pp.~1--4, IEEE, 2019.

\bibitem{closedloop}
M.~B. Milde, H.~Blum, A.~Dietm{\"u}ller, D.~Sumislawska, J.~Conradt,
  G.~Indiveri, and Y.~Sandamirskaya, ``Obstacle avoidance and target
  acquisition for robot navigation using a mixed signal analog/digital
  neuromorphic processing system,'' {\em Frontiers in neurorobotics}, vol.~11,
  p.~28, 2017.

\bibitem{imitation}
J.~Kaiser, S.~Melbaum, J.~C.~V. Tieck, A.~Roennau, M.~V. Butz, and R.~Dillmann,
  ``Learning to reproduce visually similar movements by minimizing event-based
  prediction error,'' in {\em 2018 7th IEEE International Conference on
  Biomedical Robotics and Biomechatronics (Biorob)}, pp.~260--267, IEEE, 2018.

\end{thebibliography}

\vspace{12pt}

\clearpage

\title{Training Robots with Event-Based Vision - Bio-inspired Imitation Learning for Training Central Pattern Generator using Event-Based Vision Sensor and Spiking Neural Network\\
{\footnotesize \textsuperscript{*}Note: Sub-titles are not captured in Xplore and
should not be used}
\thanks{Identify applicable funding agency here. If none, delete this.}
}

\author{\IEEEauthorblockN{1\textsuperscript{st} Ting}
\IEEEauthorblockA{\textit{Integrated Circuits and Systems Research Lab} \\
\textit{Georgia Institute of Technology}\\
Atlanta, GA \\
jting31@gatech.edu}
}
\end{document}